\documentclass[10pt,twocolumn]{article} 
\usepackage{simpleConference}
\usepackage{amssymb}
\usepackage{latexsym}
\usepackage{amsmath}
\usepackage{amssymb}
\usepackage{latexsym}

\usepackage{url}
\usepackage{xcolor}
\definecolor{newcolor}{rgb}{.8,.349,.1}

\usepackage{rotating}
\usepackage{pdflscape}

\usepackage{amsmath,amssymb,amsfonts}
\usepackage{graphicx}
\usepackage{textcomp}

\usepackage[utf8]{inputenc} 
\usepackage[T1]{fontenc}    
\usepackage{hyperref}       
\usepackage{url}            
\usepackage{booktabs}       
\usepackage{amsfonts}       
\usepackage{nicefrac}       
\usepackage{microtype}      
\usepackage{lipsum}
\usepackage{framed,multirow}
\usepackage{amssymb}
\usepackage{latexsym}
\usepackage{makecell}
\usepackage{url}
\usepackage{xcolor}
\definecolor{newcolor}{rgb}{.8,.349,.1}
\usepackage{ragged2e}
\usepackage{color,soul}
\usepackage{hyperref}
\usepackage{graphicx}
\usepackage{caption}
\usepackage{subcaption}
\usepackage{float}
\usepackage{color}
\usepackage{algorithm}
\usepackage[noend]{algpseudocode}

\usepackage{array}
\newcolumntype{P}[1]{>{\centering\arraybackslash}p{#1}}

\usepackage{hyperref}
\usepackage{longtable}

\begin{document}

\title{SketchCleanNet - A deep learning approach to the enhancement and correction of query sketches for a 3D CAD model retrieval system}

\author{\textbf{Bharadwaj Manda$^1$, Prasad Kendre$^1$, Subhrajit Dey$^2$, Ramanathan Muthuganapathy$^1$} \\
\small{$^1$ Indian Institute of Technology Madras $^2$ Jadavpur University}}

\maketitle
\thispagestyle{empty}

\begin{abstract}
Search and retrieval remains a major research topic in several domains, including computer graphics, computer vision, engineering design, etc. A search engine requires primarily an input search query and a database of items to search from. In engineering, which is the primary context of this paper, the database consists of 3D CAD models, such as washers, pistons, connecting rods, etc. A query from a user is typically in the form of a sketch, which attempts to capture the details of a 3D model. However, sketches have certain typical defects such as gaps, over-drawn portions (multi-strokes), etc. Since the retrieved results are only as good as the input query, sketches need cleaning-up and enhancement for better retrieval results.

In this paper, a deep learning approach is proposed to improve or clean the query sketches. Initially, sketches from various categories are analysed in order to understand the many possible defects that may occur. A dataset of cleaned-up or enhanced query sketches is then created based on an understanding of these defects. Consequently, an end-to-end training of a deep neural network is carried out in order to provide a mapping between the defective and the clean sketches. This network takes the defective query sketch as the input and generates a clean or an enhanced query sketch. \textcolor{black}{Qualitative and quantitative comparisons of the proposed approach with other} state-of-the-art techniques show that the proposed approach is effective. \textcolor{black}{The results of the search engine are reported using both the defective and enhanced query sketches, and it is shown that using the enhanced query sketches from the developed approach yields improved search results.}
\end{abstract}

\textit{\textbf{Keywords}} - Search and Retrieval, Engineering Models, Sketch enhancement, Mechanical Components, CAD models, Sketch-based retrieval

\section{Introduction}
\label{sec1}
The use of search engines has become commonplace in daily life. The primary driver for developing efficient search engines is the explosion of publicly available data, combined with the user's need to find relevant information from a large collection of data items \cite{baeza1999modern}. This has led to a rapid growth in the development and usage of search engines across a multitude of application domains, including library management systems, demographic data, internet-based information retrieval, enterprise search, and so on. The user first characterises his/her information need in the form of an input query and expects to find information that is most relevant to the query \cite{schutze2008introduction}. Traditionally, input queries to the search engine have been in the form of text (string of characters). This has been an effective way for users to provide their queries, since most retrieval systems focused on text data such as documents, personnel records, news articles, etc. 

However, the process of search and retrieval becomes challenging in the domain of visual data, especially 3D models, because it is difficult to characterise what a human being sees and perceives in the form of a text query \cite{textquery1}. As a result, content-based retrieval systems have been developed, which make use of the visual/shape properties of the 3D models as opposed to the traditional text query \cite{1314502, 4267943}. Among the available query options, a sketch-based query is the most efficient since it is natural for the user and easy to learn \cite{2386318}. The usage of a sketch-based query is also  shown to be very intuitive and convenient for the user than describing the 3D object by a set of descriptor rules or using the 3D model itself \cite{1268530, 6411819}.
\newline
The increased availability of large-scale datasets, paired with greater computational capabilities, has accelerated the deployment of deep learning solutions to a wide range of research problems, \textcolor{black}{including three-dimensional shape search (3DSS), which is a key problem in the engineering design process. The Princeton 3D model search engine \cite{:2004:A3M}, PROBADO3D search engine \cite{blumel2010probado3d} and the CAD model search engine proposed by \cite{li2020computer} are all well-known examples of 3DSS.} \cite{Eitz} contributed the earliest benchmark dataset of sketches corresponding to each 3D model in the Princeton Shape Benchmark (PSB) \cite{psb}. Consequently, a few instances of the Shape Retrieval Contest (SHREC) such as \cite{SHREC13} and \cite{SHREC14} established large-scale benchmarks for sketch data for 3D shapes, in addition to many approaches for sketch-based retrieval. These datasets and approaches, however, are solely aimed at generic 3D shape data or graphical representations and do not include engineering CAD model data. 

\textcolor{black}{Collecting the right information for product design is a time-consuming process. The product development process in the modern day is fast-paced, in order to keep up with rising customer demands. As a result, the development of new product designs is increasingly reliant on reference products or existing designs \cite{collaborativedesign, Funkhouser:2004:ME:1015706.1015775, albers2017agile}.  Consequently, efficient search and retrieval of relevant product designs becomes crucial for tasks such as automated feature recognition of CAD models \cite{neb2020development}, extracting components for CAD assembly \cite{lupinetti2016automatic}, computing geometric similarity of CAD parts \cite{bickel2021comparing} and so forth.}

\textcolor{black}{In product life cycle (PLC) analysis and management, the entire cycle starts with design. A user's intent of the design is best captured using a sketch of the model than a user drawing a 3D model. It is also perhaps easier to hand sketch the design intent than to use a 2D drawing tool. Once a sketch is drawn, its corresponding 3D model may be retrieved using a search and retrieval of CAD models. However, it is always not an easy task to provide a perfect sketch and often it needs cleanup to be used in further downstream applications. }

\begin{figure}[t]
    \centering
    \begin{subfigure}[b]{0.2\textwidth}
        \centering
        \includegraphics[scale=0.15]{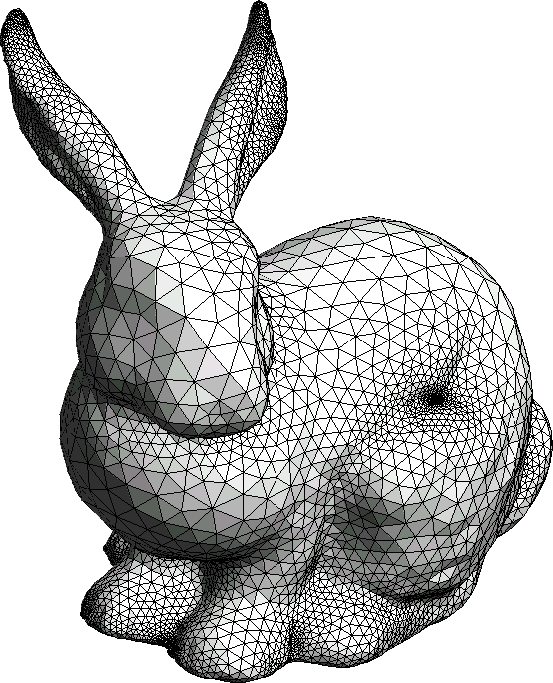}
        \caption{Generic 3D Shapes  or Graphical Models}
        \label{fig1a}
    \end{subfigure} \qquad
    \begin{subfigure}[b]{0.2\textwidth}
        \centering
        \includegraphics[height=3.5cm,width=4cm]{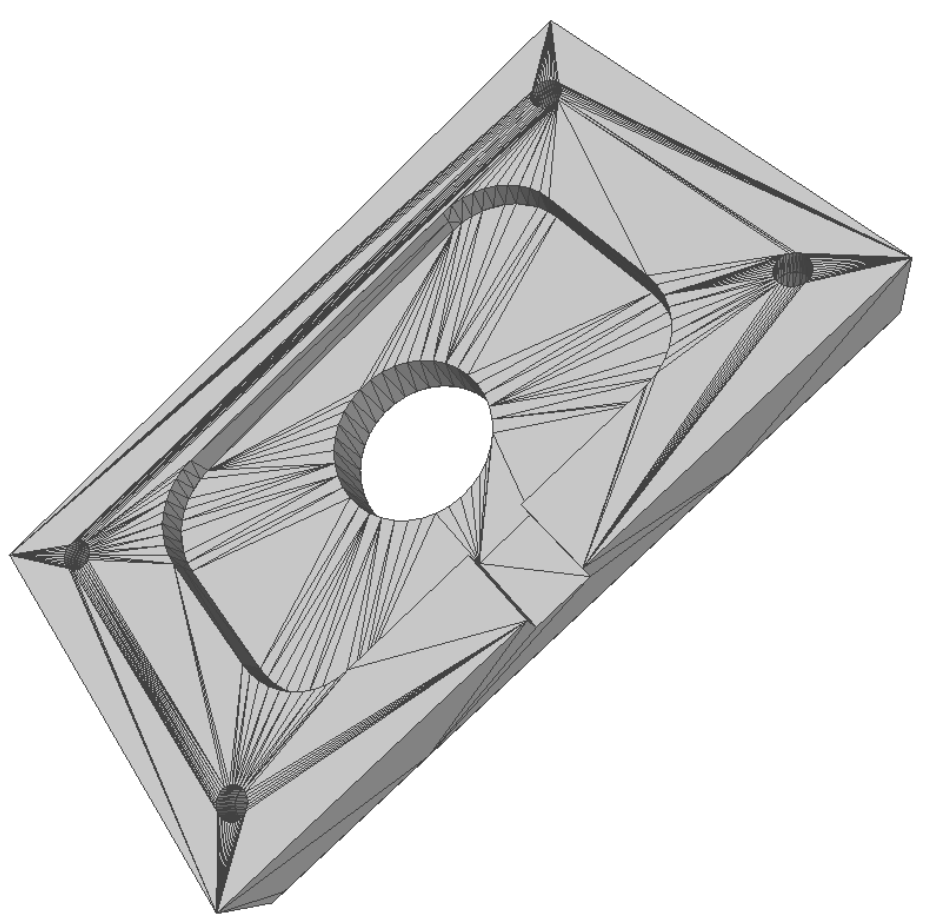}
        \caption{3D Engineering shapes or CAD models}
        \label{fig1b}
    \end{subfigure}
    \caption{Distinction between a Graphical Model and a CAD Model}
\end{figure}

\textcolor{black}{Despite the technological advances,} the progress in the \textcolor{black}{field of developing efficient search and retrieval systems for 3D CAD models} has been slow. This is largely attributed to the inherent challenges associated with CAD models of engineering parts and shapes, as well as the proprietary nature of such CAD datasets \cite{Qin2014}. An engineering components database typically includes 3D CAD models such as washers, pistons, connecting rods, gear parts, and so forth. Figures \ref{fig1a} and \ref{fig1b} illustrate a generic 3D shape and an engineering CAD model, respectively. Engineering CAD models are distinct from regular 3D shapes in the following ways \cite{ESB}:

\begin{itemize}
    \item CAD model tessellations are typically sparse
    \item CAD models contain sharp changes in curvature while generic 3D shapes are usually smooth
    \item Machining features such as holes, pockets and slots are present in 3D engineering shapes
\end{itemize}

\begin{figure*}
\centering
\includegraphics[scale=0.24]{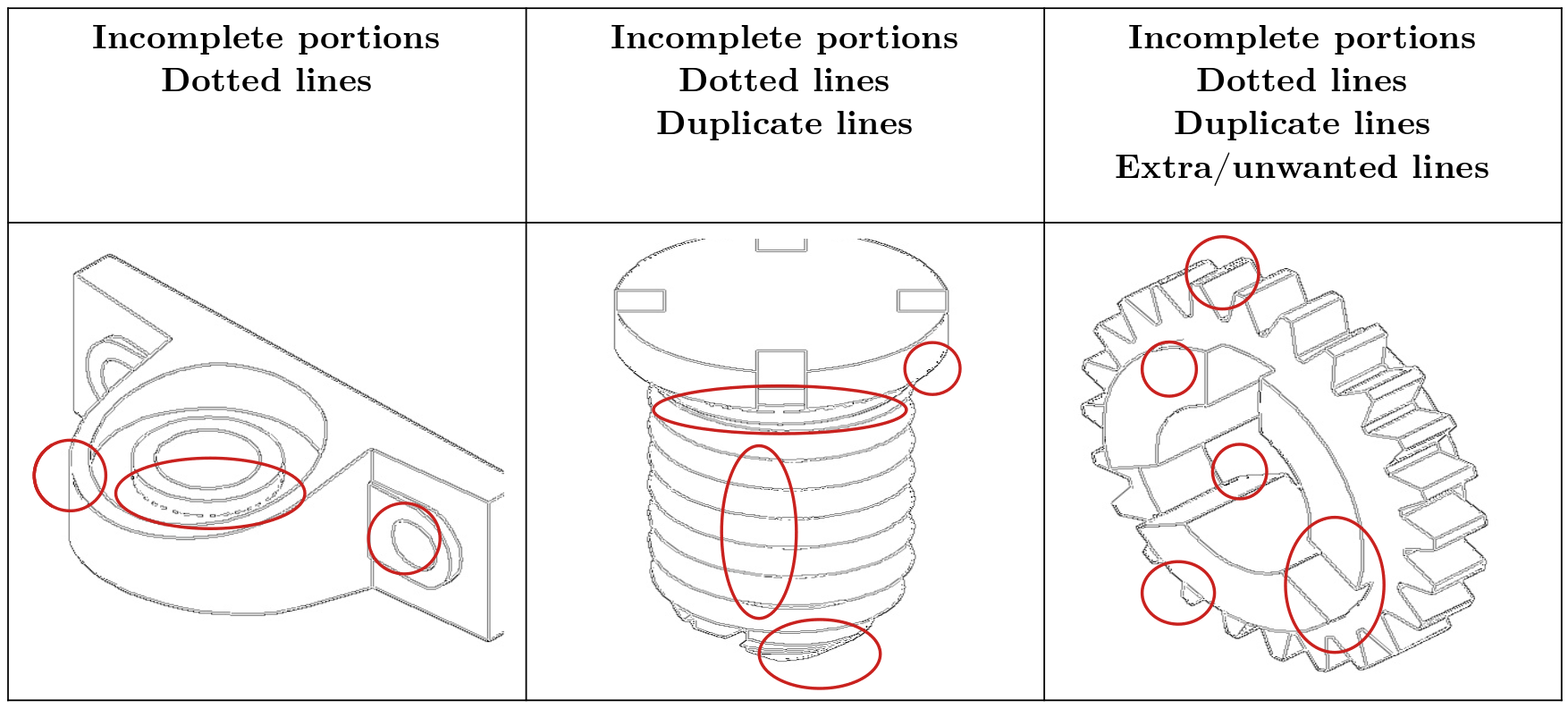}
\caption{Illustrating the various types of defects that could be present in a query sketch.}
\label{fig:defects}
\end{figure*}

In the case of engineering models, the individuals involved must have substantial domain experience and skill. Hence, the available sketch datasets for engineering CAD models are very few. The CADSketchNet dataset \cite{CADSketchNet} introduced a benchmark sketch dataset for engineering components. Query sketches for each CAD model from the Engineering Shape Benchmark (ESB) \cite{ESB} as well as the Mechanical Components Benchmark (MCB) \cite{mcb} have been provided. However, sketches have certain typical defects such as missing lines, over-drawn portions (multi-strokes), etc. \textcolor{black}{Such sketches are referred to as rough sketches or defect sketches in this paper (see Figure \ref{fig:defects} for examples).} Since the retrieved results of a search engine are only as good as the input query, these sketches need cleaning-up and enhancement for better retrieval results. \textcolor{black}{Henceforth, in this paper, "clean sketches" refers to those sketches in which such defects are either absent or greatly reduced. It also refers to the increased visual appeal of the images as compared to the "defect sketches".} It may be noted that the sketches provided in CADSketchNet \cite{CADSketchNet} contain certain defects and do not provide any clean sketches as ground truth. Hence, there is a need to first create a database of training examples that can be used to train a deep learning model. It was also established in CADSketchNet that  CAD models need separate attention as opposed to using the techniques developed for regular (also sometimes termed as graphical) models.

\textcolor{black}{This paper, therefore, presents a deep learning approach to clean or enhance the query sketches.} A dataset of enhanced query sketches \textcolor{black}{is first created} based on an understanding of the possible defects in the query sketches. \textcolor{black}{Following this, a deep neural network architecture (SketchCleanNet) is proposed, that processes the input query sketch and generates a clean/enhanced query sketch.} An end-to-end training of SketchCleanNet is carried out in order to provide a mapping between the defective and clean sketches. The proposed approach is also compared, \textcolor{black}{both qualitatively and quantitatively}, with other state-of-the-art techniques, and is shown to be more effective. \textcolor{black}{Using the enhanced query sketches yields better search results than using the query sketches with defects, as shown in Figure \ref{fig:defects}.} The key contributions of the paper are:

\begin{itemize}
    \item The first learning-based strategy to clean the rough query sketches of 3D CAD models, to the best of the authors knowledge
    \item Introduces SketchCleanNet - an end-to-end image translation scheme to understand the mapping between rough sketches and clean query images
    \item \textcolor{black}{A novel scheme to calculate the loss based on a weighted combination of pixel probabilities using Kernel Density Estimation and Weighted Cross entropy Loss}
    \item Dataset Contribution: The resulting enhanced query sketch dataset will be made available publicly 
\end{itemize}

The paper is organized as follows: Section \ref{sec:lit} discusses the related works. The dataset preparation is presented in Section \ref{sec:data} followed by the proposed methodology in Section \ref{sec:method}. A report of the results followed by a discussion on the state-of-the-art comparison is provided in Section \ref{sec:result}. Section \ref{sec:limi} discusses the potential limitations and future work, followed by a conclusion (Section \ref{sec:conc}).

\section{Related Works}
\label{sec:lit}
There have been very few studies that have focused on sketch cleanup and enhancement as a research topic in and of itself. There is little available research on the enhancement of query sketches of 3D CAD models, not to mention the available literature on learning-based approaches to such a problem. As a result, related works pertaining to algorithm-based sketch cleanup of image data and generic 3D shapes are discussed. Additionally, a few deep learning based edge detection techniques are also discussed since they aim at generating enhanced sketches from the input image(s). A summary of the available 3D CAD model datasets is also presented.

\subsection{Algorithms for sketch simplification}
\cite{barla} were one of the first to present an algorithm for cleaning rough sketches. An input vector drawing is simplified by clustering input strokes based on the geometry, replacing each cluster with a single representative line/curve. Consequently, many works such as \cite{Liu2015, Liu2018, 7899777} were proposed, that follow a similar line of thought by classifying or clustering the input strokes and replacing each cluster with a single line/curve. The work presented by \cite{Noris} focused on simplifying raster inputs. \cite{parakkat2018delaunay, donati2019complete} also focus on cleaning rough raster sketches using Delaunay triangulation and Bézier splines respectively. \cite{Yan-benchmark} provides a benchmark for rough sketch cleanup by comparing many state-of-the-art techniques. 

\cite{Simo-FCN} have introduced a neural network based approach to cleaning rough sketches, where a series of convolutional layers are used. A new dataset that contains 68 pairs of rough and simplified sketches (obtained from 5 artists) is used to train the Fully Convolutional Network. The authors attempt to better their sketch simplification by proposing an unsupervised Generative Adversarial Network (GAN) for sketch simplification \cite{Simo-master}, in an attempt to overcome the limited availability of annotated training pairs in their previous work. All these methods, however, only aim at simplifying the sketches of generic shapes and objects, and do not focus on the 3D shapes of engineering parts and components.

\subsection{Edge detection techniques}
Edge-detection techniques have been traditionally used for several computer vision tasks. A survey of classical edge detectors is provided in \cite{ziou1998edge}. With the advent of neural networks, many sophisticated edge detection techniques have been proposed. An effective edge detection procedure could potentially serve as a sketch generation algorithm. Holistically Nested Edge Detection (HED) was one of the earlier methods to use a neural network approach for an end-to-end edge detection system \cite{HNED2015}. DeepEdge \cite{DeepEdge2015} proposes a multi-scale bifurcated deep network, consisting of two independently trained branches. While one branch attempts to detect contours, the second branch is optimized to depict the fraction of human annotators agreeing to the presence of a contour. 

\cite{BDCN2019} have introduced a Scale Enhancement Module to their convolutional neural net (CNN), that effectively increases the size of the receptive fields of network neurons. \cite{DenseXtNet2020} attempt to build a robust CNN model (DexiNed) for edge detection, inspired by both HED and Xception \cite{Xception} networks. A large dataset with edge annotations is also introduced here. However, the dataset consists only of objects from the generic object categories. \cite{neural_contours} proposed the idea of neural contours, wherein a neural network is trained to generate line drawings of 3D models. The proposed approach yields state-of-the-art drawings, owing to its complex neural network pipeline, but at a heavy computational cost. 

\subsection{Available Sketch datasets of common objects}
\cite{Eitz} provided one of the first sketch datasets, containing hand-drawn sketches corresponding to every 3D model in the Princeton Shape Benchmark \cite{psb}. A few instances of the Shape Retrieval Contest (SHREC) have also provided benchmark sketch datasets \cite{SHREC13, SHREC14}. The QuickDraw dataset \cite{quickdraw} provides a collection of vector drawings from an online game, where users have to provide rough sketches in less than 20 seconds. The OpenSketch dataset \cite{OpenSketch19} contains annotated product design sketches, including a detailed study of stroke pressure and drawing time etc. The dataset, however, contains only a limited number of sketches (107) across 12 categories. The ProSketch3D \cite{ProSketch} consists of 1500 sketches of 3D models across 500 object categories, taken from ShapeNet \cite{ShapeNet}. 

\subsection{Datasets of 3D CAD models}
The Engineering Shape Benchmark (ESB) \cite{ESB} was one of the earlier datasets of 3D CAD models, with 801 CAD models across 42 classes. The CADNET dataset \cite{CADNET} combines the ESB dataset and the National Design Repository (NDR) \cite{NDR} and augments them with additional CAD models, resulting in 3317 CAD models across 43 classes. \cite{ABC} have introduced the ABC dataset with a million CAD models. \textcolor{black}{However, it is difficult to make use of this dataset for tasks such as automated classification and retrieval, because the ground truth information such as category labels is not available in the dataset.} The Mechanical Components Benchmark (MCB) \cite{mcb} is the latest benchmark dataset of 3D CAD models, containing 58,696 CAD models across 68 categories. These datasets only provide the 3D CAD model data and do not contain any sketch information.

A sketch dataset for 3D CAD models is proposed by \cite{Qin2017}, consisting of 2148 CAD models and six corresponding views of each model. However, this dataset is proprietary and is not available. \cite{SketchGraphs} introduce SketchGraphs, a sketch-dataset for CAD models based on the design workflow as opposed to the model shape and geometry. These datasets are not useful in developing a deep learning based search engine for 3D CAD models. The recently proposed CADSketchNet dataset \cite{CADSketchNet}, contains 801 hand-drawn sketches for every CAD model in the ESB. Additionally, a weighted combination of the Canny edge detection algorithm and Gaussian Blurring is proposed as a sketch-generation algorithm, which is used to obtain 58,696 computer-generated sketches for every CAD model in the MCB. This dataset is both large-scale and well-annotated, making it a suitable choice for developing deep learning based solutions.

\section{Dataset Preparation}
\label{sec:data}
Given a rough sketch of a 3D CAD model as an input, the goal of this research is to build a deep learning model that can generate clean or enhanced query sketches. The dataset needed to train such a deep learning model, should contain training pairs as follows: ($X$, $Y$) - where $X$ is the rough sketch of the 3D CAD model (which is to be cleaned), and $Y$ is the clean image (ground truth). A significant number of such training examples are needed, in order for the deep learning model to effectively learn the mapping between the images. In the absence of any such dataset, a new dataset is to be constructed from scratch. 

For this purpose, the computer-generated sketches from the CADSketchNet dataset are used as the rough sketches ($X$). While the proposed sketch-generation algorithm in CADSketchNet \cite{CADSketchNet} is automatically able to produce computer-generated query sketches for 3D CAD models, the resulting sketches have a few defects. A brief overview of various sketches in the dataset indicates the presence of certain defect types, such as missing partial input lines, presence of duplicate lines, unwanted mesh lines in sketches, extra lines that are not present in input etc. Since the retrieved results of a search engine are only as good as the input query, these sketches need cleaning-up or enhancement in order to obtain better retrieval results. These sketches, therefore, serve as excellent candidate images for $X$.

\begin{figure*}
\centering
\includegraphics[width=1\textwidth,height=10cm]{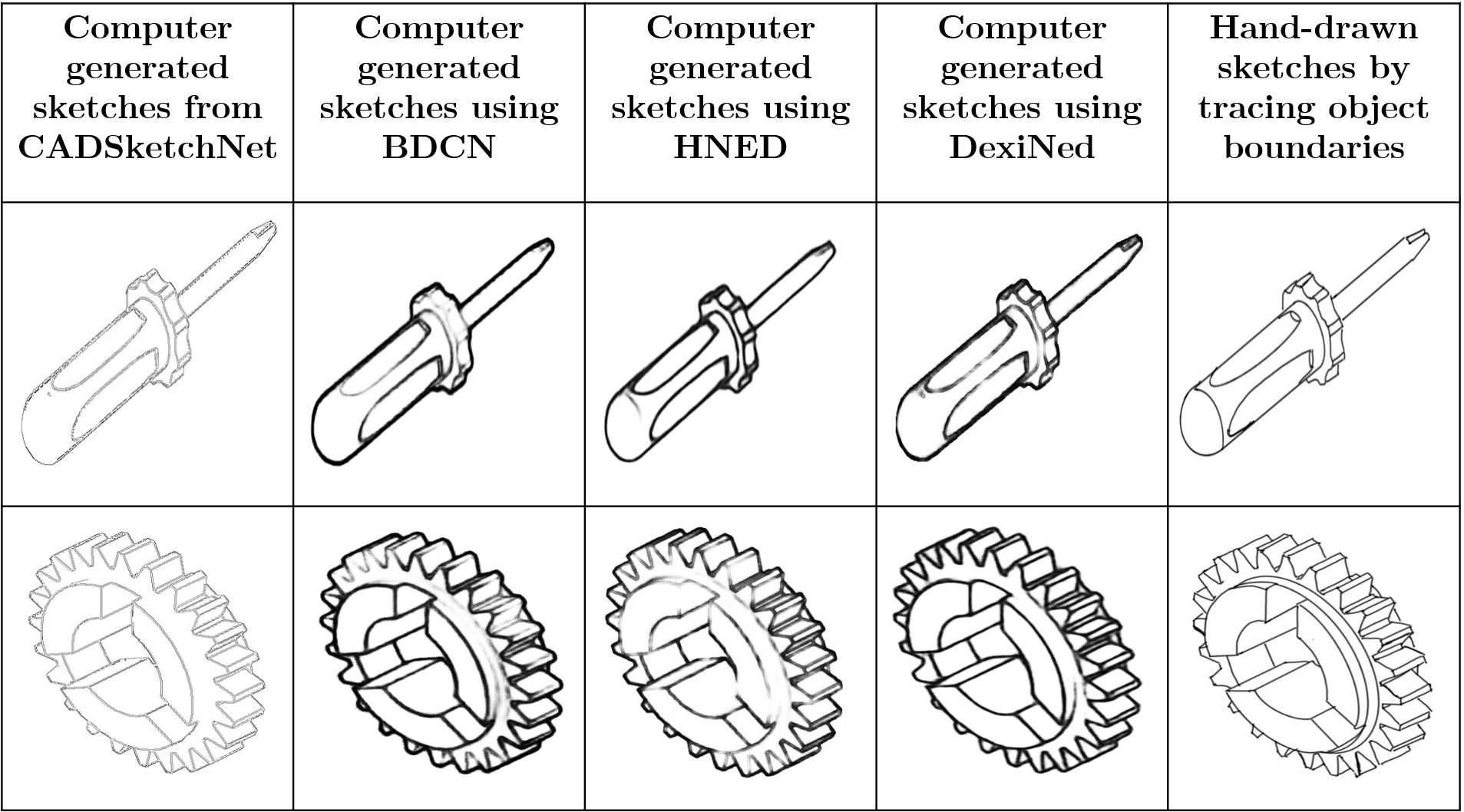}
\caption{Hand-drawn sketches by tracing object boundaries provide the best ground truth for training. Bi-Directional Cascade Network (BDCN) for edge detection \cite{BDCN}; HNED - Holistically-Nested Edge Detection \cite{HNED}; DexiNed - Dense Extreme InceptionNet \cite{DenseXNet}}
\label{fig:data_pairs}
\end{figure*}

The ground truth $Y$ (clean images) of the corresponding sketches are not available in the dataset. Edge detection techniques result in images that can potentially be used as sketches. Hence, some of the popular neural network based edge detectors are experimented with, to obtain potential candidate images for $Y$. After due experimentation, it has been found that none of the networks are able to provide clean sketch images (see Figure \ref{fig:data_pairs}). Hence, we manually generate training pairs by hand-drawing sketches by tracing the object boundaries for every CAD model. This method results in clean and sharp sketch images, which are then used as the ground truth image $Y$. It is important to note that the `traced' sketches are obtained only to be used as ground truth images for training and not as a substitute for the query images provided to the search engine. \textcolor{black}{It should also be noted that the ground truth is obtained merely by tracing the object boundaries from the CAD model images and not "drawing" new sketches. Additionally, it is made sure that the obtained ground truth are free from potential noise and errors by validating them by a taking a majority vote among a group of volunteers with knowledge of CAD and drawing. This avoids the potential issue of worsening the input sketch by introducing an intermediate sketch enhancement module.}

Since it is not possible to manually obtain the `traced' sketches for a dataset as large as the MCB, the training pairs are generated only for the 801 CAD models in the ESB. \textcolor{black}{It should be noted here that the $X$ here contains 801 computer-generated sketches of the ESB, using the sketch-generation algorithm proposed by \cite{CADSketchNet}. These are not the 801 hand-drawn sketches from CADSketchNet. As discussed earlier in this section, although \cite{CADSketchNet} proposed an algorithm to automatically generate query sketches for 3D CAD models, the resulting sketches have a few defects. Therefore, we use these defective query images of the ESB as the candidate images for $X$ and the corresponding traced sketches as the clean images for $Y $(ground truth).} \textcolor{black}{This dataset with 801 pairs is split into 632 (for the train set) and 169 (for the test set) respectively. This is the standard 80-20 split ratio as per the Pareto principle.}

\section{Network Architecture}
\label{sec:method}
The problem at hand involves an end-to-end training, an attempt to map the relationship between the two images - the rough sketch $X$ and the clean sketch $Y$. A traditional fully connected network cannot be used since it does not handle image inputs. A convolutional neural network (CNN) can process image inputs, but the fully connected layers at the end are meant for classification tasks. The suitable choice of network architecture for an end-to-end image translation problem would be the use of a Fully Convolutional Network (FCN). The characteristic features of an FCN are: (1) the absence of any fully connected layers; (2) the use of up-sampling or de-convolution layers; and (3) the presence of an encoder-decoder network, resulting in an hourglass-shaped architecture. 

\begin{figure*}
\centering
\includegraphics[width=1\textwidth,height=8cm]{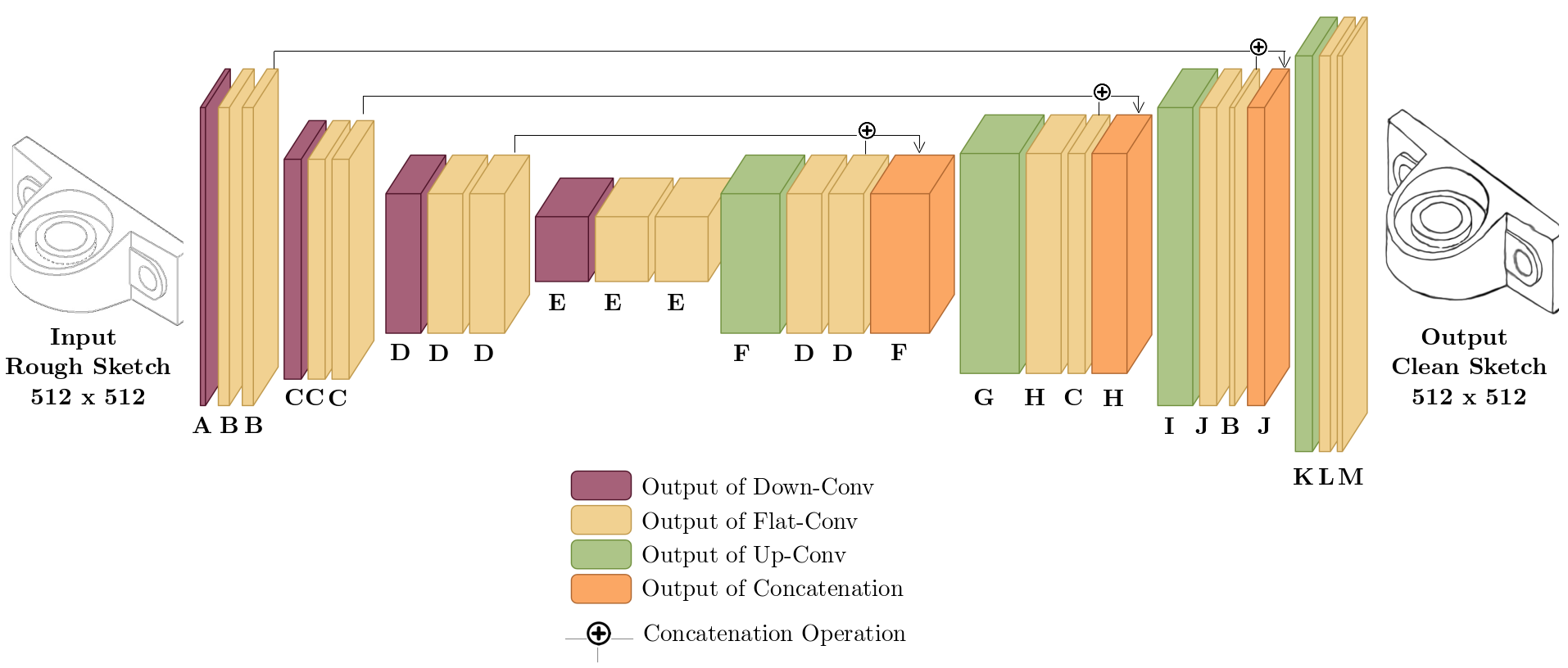}
\caption{Proposed Fully Convolutional Network Architecture called `SketchCleanNet'. The dimensions at every layer are \textbf{A:} 32 x 256 x 256; \textbf{B}: 64 x 256 x 256; \textbf{C:} 128 x 128 x 128; \textbf{D:} 256 x 64 x 64; \textbf{E:} 512 x 32 x 32; \textbf{F:} 512 x 64 x 64; \textbf{G:} 512 x 128 x 128; \textbf{H:} 256 x 128 x 128; \textbf{I:} 256 x 256 x 256; \textbf{J:} 128 x 256 x 256; \textbf{K:} 128 x 512 x 512; \textbf{L:} 64 x 512 x 512; \textbf{M:} 32 x 512 x 512}
\label{fig:fcn}
\end{figure*}

The proposed network architecture for the current task is shown in Figure \ref{fig:fcn}. We name this network `SketchCleanNet' (referred to as SCNet henceforth). The network consists of four types of operations: down-convolution, flat-convolution, up-convolution, and concatenation using skip connections. Down-convolution is the standard convolution operation that decreases the input size by a pre-determined factor. The flat-convolution operation performs the standard convolution operation while maintaining the input size. Up-convolution operations are used in the decoder portion of the network, which merely attempts to up-sample the data to a higher dimension. The network also contains skip connections, which are mainly used to avoid the issue of vanishing gradient \textcolor{black}{and to aid the reconstruction of input structure in the output images}. Concatenating a few layers in the decoder part with skip connections from the encoder portion aids in reconstructing the image to the original dimension. \textcolor{black}{Table \ref{tab:ablation2} shows the effect of skip connections on the network performance.}

\renewcommand{\arraystretch}{1.50}
\begin{table}[t]
\centering
\resizebox{0.47\textwidth}{!}{%
\begin{tabular}{|c|c|c|c|c|c|}
\hline
& \textbf{MSE ↓}    & \textbf{L1 ↓} & \textbf{BDCN Loss ↓} & \textbf{PSNR ↑}  & \textbf{SSIM ↑}  \\ \hline
\textbf{With skip connections}    & \textbf{0.0149} & \textbf{0.026} & \textbf{0.9024}  & \textbf{18.86} & \textbf{0.9208} \\ \hline
\textbf{Without skip connection} & 0.0171          & 0.0281         & 0.9827              & 18.29          & 0.9101          \\ \hline
\end{tabular}%
}
\caption{Comparing The effect of skip connections on the network performance across different metrics. Using skip connections helps to obtain a better reconstruction of the input image. MSE - Mean Squared Error; BDCN Loss - \cite{BDCN2019}; PSNR - Peak Signal-to-Noise Ratio; SSIM - Structured Similarity Index \cite{SSIM}; ↑ - greater value for the metric indicates higher similarity; ↓ - lesser value for the metric indicates higher similarity;}
\label{tab:ablation2}
\end{table}
\renewcommand{\arraystretch}{1.00}

The key highlights of the proposed network architecture are:
\begin{itemize}
    \item The absence of pooling layers: Ideally, a pooling operation is expected to extract only useful information and discard irrelevant details. However, this might not be the case every time. There is always a scope for losing useful information. For the current scenario, since the fraction of pixels in the input sketch that contain lines/curves is already small, the pooling layers are not used in the proposed network. This has been verified by experimentation, wherein the network that contained pooling layers could not provide an accurate reconstruction of the input image.
    
    \item Substituting the use of pooling layers with down-convolution layers: Convolution layers contain trainable parameters as opposed to a pooling layer. Therefore, the advantages of substituting the pooling operation with down-convolution layers are two-fold: (1) No information is lost as a result of pooling; (2) Aids in training the network to obtain a better mapping between input output images.
    
    \item The use of flat-convolution layers to avoid shrinking the image dimension each time a filter is applied from the convolution operation. This is mainly done to keep the network deep. Otherwise, the input image dimensions would quickly reduce, which drastically reduces the number of convolution layers that can be used in the network architecture.
\end{itemize}

\subsection{Implementation Details}
The current choice of network architecture for SCNet (SCNet) is made after due experimentation with respect to the number of layers, kernel sizes for convolutions, and so forth. The choices of such hyper-parameters are based on heuristics and practical training strategies such as \cite{Goodfellow, Bengio2012PracticalRF}. The network is trained \textcolor{black}{(on the training set that contains 632 training pairs)} for 200 epochs, and uses the Adam optimization algorithm \cite{Adam}. A learning rate of 3$e$-04 and a batch size of 8 are used. ReLU is used for activation in all layers. A kernel size of (3,3) is used in all convolutional layers. The stride for every down-convolution layer is (2,2) with padding = 1; the flat-convolution layer is (1,1) with padding = 1. A scale of 2 is used in each up-convolution layer. 

\subsection{Choice of Loss Function}
While the standard cross-entropy loss function is typically used for image classification tasks, it does not yield good results in the current scenario. This is because the distribution of edge/non-edge pixels in a typical sketch image is substantially skewed - at least 90\% of the ground truth is white pixels. The Bi-Directional Cascade Network (BDCN) for edge detection \cite{BDCN} uses a simple strategy to counter this, which essentially calculates the number of positive and negative pixels (with edge and without edge). Every pixel that has a value below a certain threshold is considered as a positive pixel \textcolor{black}{($Y_{+}$)} and other pixels are considered negative pixels \textcolor{black}{($Y_{-}$)}. The consequent loss function is defined as:
\begin{equation} \label{eq1}
\mathcal{L}_1=-\alpha \sum_{j \in Y_{-}} \log \left(1-\hat{y}_{j}\right)-\beta \sum_{j \in Y_{+}} \log \left(\hat{y}_{j}\right)
\end{equation} 
where $\alpha=\lambda \cdot\left|Y_{+}\right| /\left(\left|Y_{+}\right|+\left|Y_{-}\right|\right)$ and $\beta = \left|Y_{-}\right| /\left(\left|Y_{+}\right|+\left|Y_{-}\right|\right)$ are the relative weights for the positive and negative pixels, with $\lambda$ being a hyper-parameter to control the weight of positive over negative samples.

DeepHist \cite{avi2020deephist} proposes the idea of a Kernel Density Estimation (KDE) function. The main idea here is to represent the image as a histogram, where the height of each bin in the histogram represents the probability of a range of pixels being present in that bin. For instance, the outputs of SCNet are single-channel sketch images which have a large number of white pixels. So the bin corresponding to the group of white pixels will be taller in the histogram, thus indicating a high probability of the presence of white pixels.

For training SCNet, we use a loss function that is similar to the BDCN loss function described in Equation \ref{eq1}, while also integrating the idea of KDE. The BDCN loss calculates the weights of positive and negative pixels based on a hyper-parameter threshold. This is substituted by the use of KDE. The kernel used for the current KDE implementation is given by,

\begin{equation}
k(z)=\frac{1}{\sqrt{2 \pi \sigma}} \times \exp \left(-\frac{z^{2}}{2 \sigma^{2}}\right),
\end{equation}
where $\sigma$ is the variance of the distribution.

The KDE function is given by,
\begin{equation}
\hat{f}(g)=\frac{1}{N} \sum_{x \in \Omega} k(I(x)-g),
\end{equation}
where $N$ is the number of pixels in the ground truth, $I(x)$ is the intensity of the pixel $x$ and $\Omega$ is the set of all pixels in the ground truth image. The method used to calculate the weights for the positive and negative pixels is as follows. All pixels of the ground truth image are in the range [0,1]. If we have $K$ bins, each bin will be of the length $\frac{1}{K}$. The probability of a pixel $g$ falling in the $k$-th bin ($B_k$) is given by,

\begin{equation}
\operatorname{P}\left(g \in B_{k}\right)=\int_{B_{k}} \hat{f}(g) \cdot d g
\end{equation}

The bin with the maximum probability ($P_{max}$) of containing the pixel $g$ is determined. This process is repeated for every pixel in the ground truth image and these probability values are then used as weights in the loss function. The expression for the loss function using Kernel Density Estimation is given by, 

\begin{equation} \label{eq2}
\mathcal{L}_2 =-\sum_{g} P_{max_g} \times y_{g} \log \left(\hat{y}_{g}\right)
\end{equation}

\begin{figure*}
\centering
\includegraphics[scale=0.45]{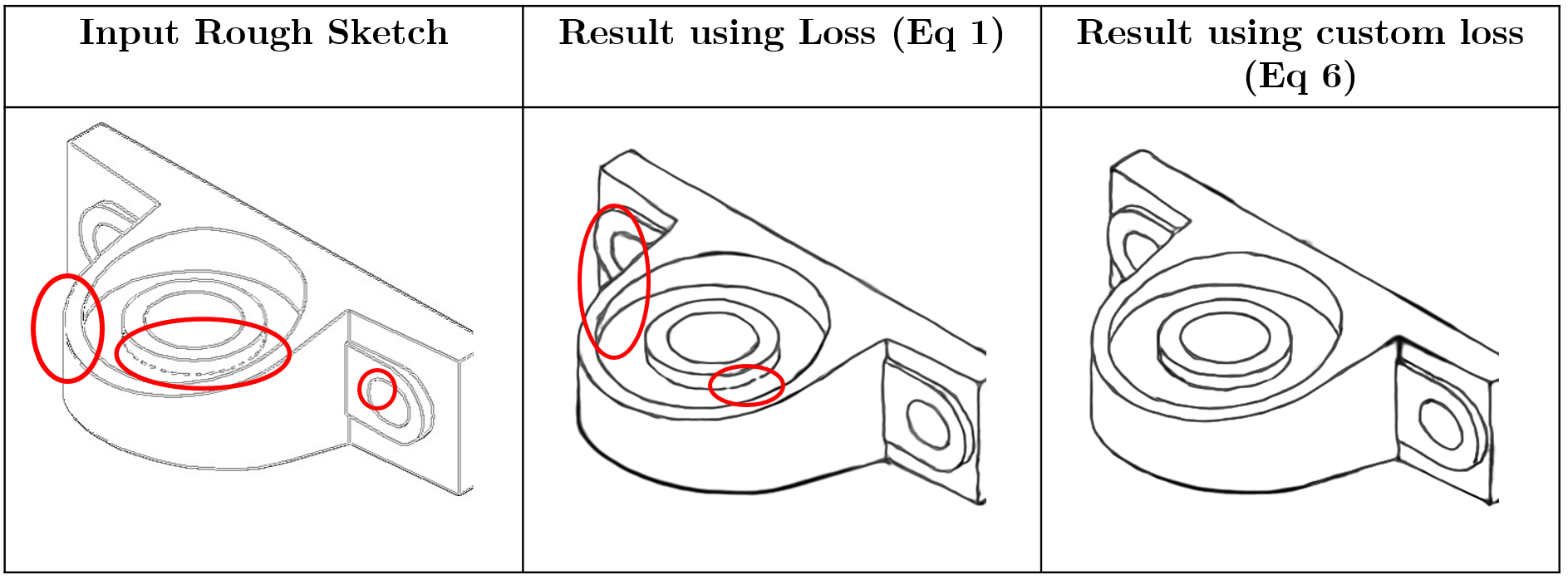}
\caption{Using the custom loss function yields cleaner and sharper sketch images. The network output using the loss from Eq \ref{eq1} still contains gaps and blurred regions, which are resolved using the proposed loss function.}
\label{fig:loss_compare}
\end{figure*}

The custom scheme to calculate losses at the final layer of the network uses a weighted combination of the BDCN loss (Equation \ref{eq1}) and the loss using KDE (Equation \ref{eq2}).

\begin{equation}
\mathcal{L} = \lambda_1 \mathcal{L}_1 + \lambda_2 \mathcal{L}_2,
\end{equation}

where $\lambda_1, \lambda_2$ are hyperparameters. The values for $\lambda_1, \lambda_2$ used in the current implementation are 0.8 and 0.2 respectively, identified after due experimentation. \textcolor{black}{Table \ref{tab:ablation1} shows the effect of $\lambda_1$ and $\lambda_2$ for a few combinations}. Using the custom loss function shows a significant improvement in results over using the BDCN loss alone (see Figure \ref{fig:loss_compare}). 

\begin{figure*}
\centering
\includegraphics[scale=0.3]{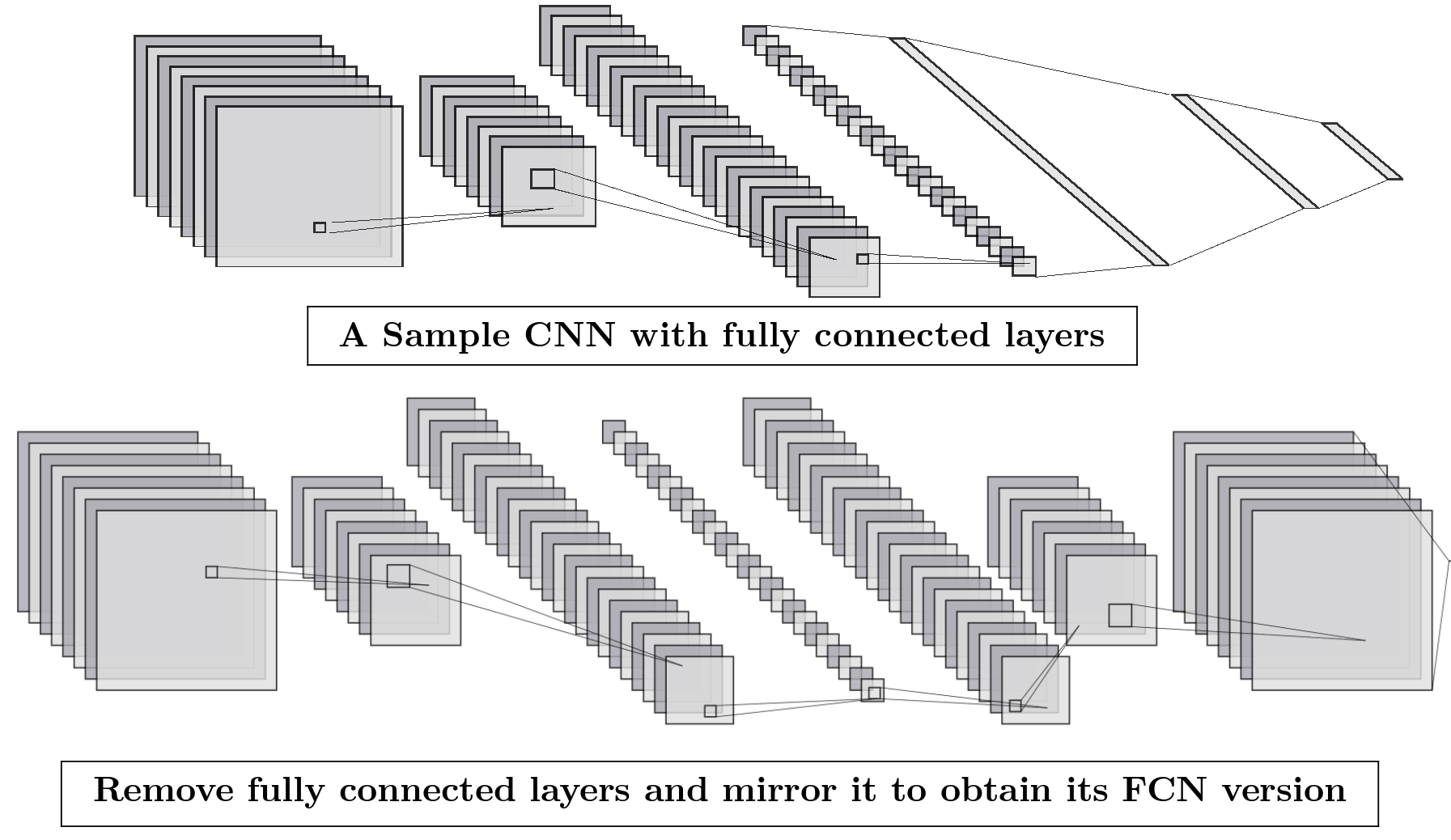}
\caption{Method used to obtain the fully convolutional version of any sample CNN with dense layers at the end.}
\label{fig:fcn_sample}
\end{figure*}

\subsection{Coding Framework and System Configuration}
For implementing the neural network, Python3 with PyTorch is used. OpenCV library is used for all image-based implementations. All the implementations are carried out on a system running Ubuntu 20.04 Operating System. The system has an Intel Core i7-8700K CPU with 64GB RAM and an NVIDIA RTX 2080Ti GPU with 12GB RAM.

\renewcommand{\arraystretch}{1.50}
\begin{table}[t]
\centering
\resizebox{0.47\textwidth}{!}{%
\begin{tabular}{|c|c|c|c|c|c|c|}
\hline
\textbf{$\lambda_1$}    & \textbf{$\lambda_1$}    & \textbf{MSE ↓}    & \textbf{L1 ↓} & \textbf{BDCN Loss ↓} & \textbf{PSNR ↑}  & \textbf{SSIM ↑}   \\ \hline
\textbf{0.8} & \textbf{0.2} & \textbf{0.0149} & 0.026       & \textbf{0.9024}     & \textbf{18.86} & \textbf{0.9208} \\ \hline
0.7  & 0.3  & 0.0167 & 0.0276          & 1.1299 & 18.37 & 0.9188 \\ \hline
0.85 & 0.15 & 0.0155 & 0.0264          & 1.0264 & 18.74 & 0.9192 \\ \hline
0.5  & 0.5  & 0.0153 & \textbf{0.0257} & 0.9689 & 18.79 & 0.9202 \\ \hline
1    & 0    & 0.0151 & 0.0258          & 0.9526 & 18.82 & 0.9199 \\ \hline
\end{tabular}%
}
\caption{The effect of various for $\lambda_1$ and $\lambda_2$. Using the values of 0.8 and 0.2 gives the best performance across different metrics.  MSE - Mean Squared Error; BDCN Loss - \cite{BDCN2019}; PSNR - Peak Signal-to-Noise Ratio; SSIM - Structured Similarity Index \cite{SSIM}; ↑ - greater value for the metric indicates higher similarity; ↓ - lesser value for the metric indicates higher similarity;}
\label{tab:ablation1}
\end{table}
\renewcommand{\arraystretch}{1.00}

\begin{sidewaysfigure*}
\centering
\includegraphics[width=1\textwidth,height=12.5cm]{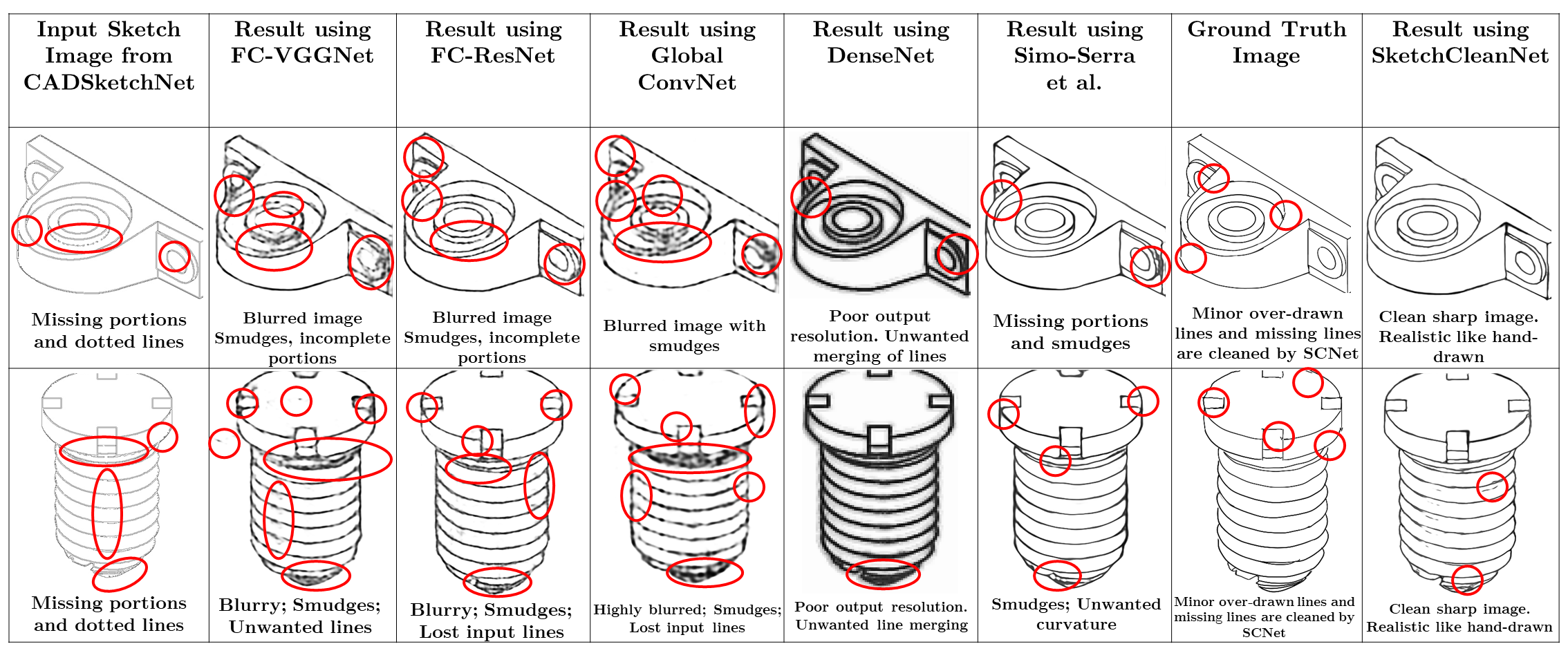}
\caption{The proposed SCNet Architecture gives the best results. FCVGGNet, FCResNet - fully convolutional versions of VGGNet \cite{VGG} and ResNet \cite{ResNet} respectively, obtained from process described in Figure \ref{fig:fcn_sample}; Global ConvNet - \cite{GCN}; DenseNet \cite{DenseNet}; Simo-Serra et al. \cite{Simo-FCN};}
\label{fig:clean_sketches}
\end{sidewaysfigure*}

\renewcommand{\arraystretch}{1.50}
\begin{table*}[t]
\centering
\begin{tabular}{|c|c|c|c|c|c|}
\hline
\textbf{Network}    &\textbf{MSE ↓} &\textbf{ L1 loss ↓} & \textbf{BDCN Loss ↓}   & \textbf{PSNR ↑}  & \textbf{SSIM ↑}   \\ \hline \hline
FC-VGGNet & 0.020 & 0.0373 & 1.5225 & 16.90 & 0.8962 \\ \hline

FC-ResNet & 0.016 & 0.0307 & 1.1245 & 17.80 & 0.9115 \\ \hline

DenseNet & 0.050 & 0.1012 & 4.4420 & 13.00 & 0.8129 \\ \hline

Simo-Serra et al. \cite{Simo-FCN} & 0.0173 & 0.0287 & 1.2270 & 17.61 & 0.9153 \\ \hline

GCN & 0.020 & 0.039 & 1.6354 & 16.62 & 0.8901 \\ \hline

\textbf{SCNet} & \textbf{0.0149} & \textbf{0.0260} & \textbf{0.9024} & \textbf{18.24} & \textbf{0.9208} \\ \hline
\end{tabular}%
\caption{Similarity results when comparing the outputs of various networks with ground truth \textcolor{black}{(on the test set)}. SCNet gives the closest results. MSE - Mean Squared Error; BDCN Loss - \cite{BDCN2019}; PSNR - Peak Signal-to-Noise Ratio; SSIM - Structured Similarity Index \cite{SSIM}; ↑ - greater value for the metric indicates higher similarity; ↓ - lesser value for the metric indicates higher similarity;}
\label{tab:compare_y}
\end{table*}

\renewcommand{\arraystretch}{1.50}
\begin{table*}[t]
\centering
\resizebox{\textwidth}{!}{%
\begin{tabular}{|c|c|c|c|c|c|c|c|}
\hline
                             & \textbf{CADSketchNet} & \textbf{VGGNet} & \textbf{ResNet} & \textbf{GCNet} & \textbf{DenseNet} & \textbf{Simo-Serra} & \textbf{SCNet} \\ \hline
\textbf{Top $k$ accuracy} & 94.23  & 92.51   & 95.79   & 92.23   & 91.51   & 96.05   & 96.43  \\ \hline
\textbf{Precision}      & 0.9375 & 0.91667 & 0.92308 & 0.87500 & 0.88889 & 0.90000 & 0.9412 \\ \hline
\textbf{Recall}         & 0.4545 & 0.3818  & 0.42222 & 0.34705 & 0.33889 & 0.43529 & 0.4705 \\ \hline
\textbf{Mean Retrieval Time} & 1.38E-05              & 1.42E-05        & 1.38E-05        & 1.49E-05       & 1.39E-05          & 1.40E-05            & 1.39E-05       \\ \hline
\end{tabular}%
}
\caption{\textcolor{black}{Clean sketches generated by SCNet give better retrieval performance when compared to the defect sketches from CADSketchNet and sketches generated by other networks.}}
\label{tab:ret_compare}
\end{table*}
\renewcommand{\arraystretch}{1.0}

\section{Results and Discussion}
\label{sec:result}
The results of the proposed SCNet are reported in this section, followed by a discussion on the comparison of results with other state-of-the-art approaches and on other sketch datasets. \textcolor{black}{All the reported results are on the test set, containing 169 images.}

\subsection{Comparing results with other networks}
Only a few state-of-the-art FCN architectures exist in the literature, such as the Global Convolutional Network (GCN) proposed by \cite{GCN}. Even so, these networks are mostly aimed towards image segmentation and not directly on sketch enhancement. Hence, we make use of a few popular ImageNet architectures, by removing the fully connected layers and mirroring them. This results in a fully convolutional version of these networks. This process is summarized in Figure \ref{fig:fcn_sample} for a sample CNN.

Some sample visualizations of the results can be seen in Figure \ref{fig:clean_sketches}. The fully convolutional version of VGGNet \cite{VGG} does not perform well and results in images that are blurred, The outputs also contain many smudges and defects of the input rough sketch are not fully removed. Similar characteristics are observed for the images generated by the fully convolutional version of ResNet \cite{ResNet}, although the images are a lot sharper. GCN performs very badly, probably because the architecture is mainly suited for image segmentation as opposed to a dedicated sketch enhancement task. It is easy to observe that the images generated by SCNet are a lot sharper with most of the input defects removed. They also give a more natural feel, bearing a closer resemblance to hand-drawn sketches.

\begin{figure*}
    \centering 
    \begin{minipage}[b]{0.49\textwidth}
        \centering
        \includegraphics[width=1\textwidth,height=5cm]{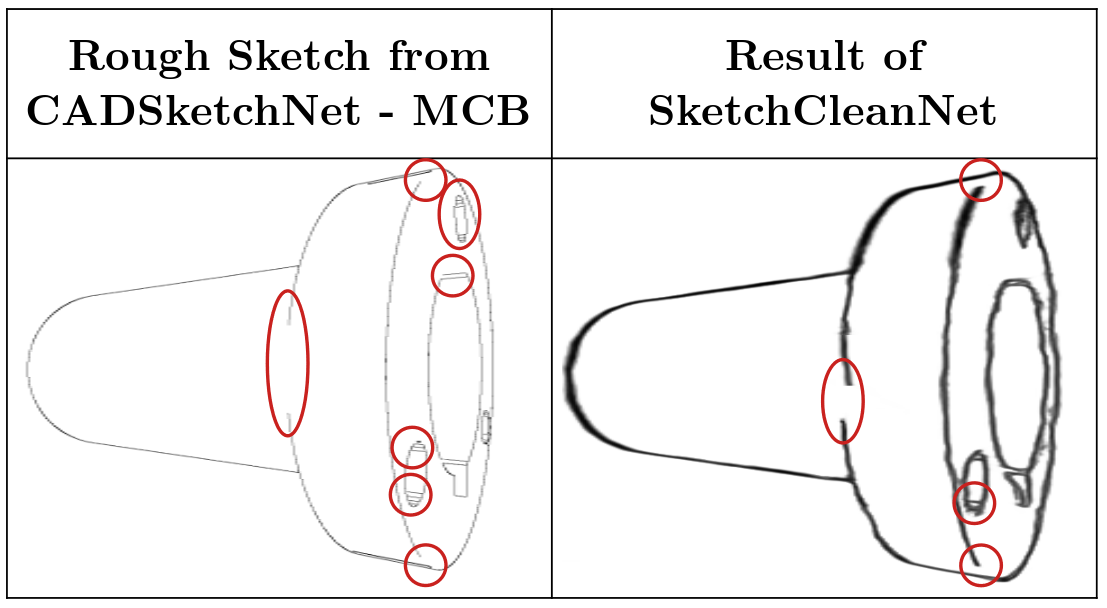}
        \caption{Result of SCNet on MCB - Sample 1. Most defects are removed from the input sketches.}
        \label{fig:result_mcb1}
    \end{minipage}
     \begin{minipage}[b]{0.49\textwidth}
        \centering
        \includegraphics[width=1\textwidth,height=5cm]{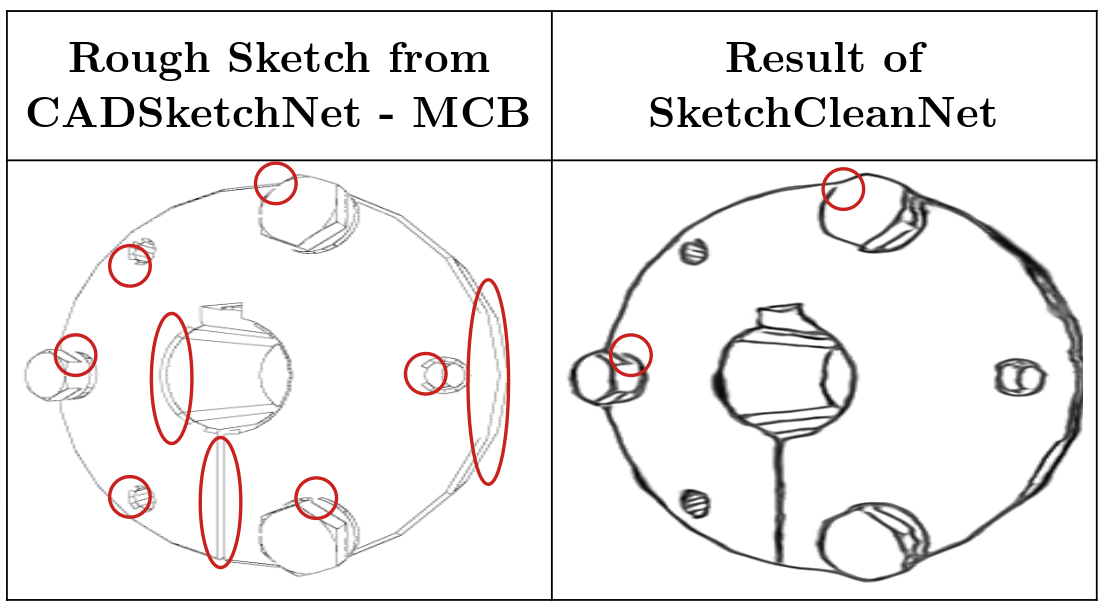}
        \caption{Result of SCNet on MCB - Sample 2.  Most defects are removed from the input sketches.}
        \label{fig:result_mcb2}
    \end{minipage}
\end{figure*}

In an attempt to quantify the visual results obtained by these methods, the output images from each of the above networks are compared for similarity with the ground truth clean sketches. Table \ref{tab:compare_y} reports the similarity values obtained by the networks using various similarity metrics. From the table, it can be seen once again that the SCNet provides the clean sketches which are closest to the ground truth for every tested metric of similarity.

\subsection{Results of SCNet on sketch datasets of CAD models}
SCNet has been trained on the rough and clean image pairs of the ESB dataset. The rough sketches of ESB were obtained from the CADSketchNet dataset. CADSketchNet also contains query sketches for every CAD model in the MCB dataset. SCNet could not be trained on these since it was difficult to obtain the ground truth clean sketches. Nonetheless, the trained SCNet can now be tested on the MCB sketches and the performance can be evaluated. A random sample of 1000 sketches of the MCB models is chosen from CADSketchNet and are fed to SCNet. A few sample results of the cleaned sketches are shown in Figures \ref{fig:result_mcb1} and \ref{fig:result_mcb2}. It can be seen that a majority of the defects are removed from the input sketches, and the network produces good results.

\begin{figure*}
\centering
\includegraphics[width=1\textwidth,height=10cm]{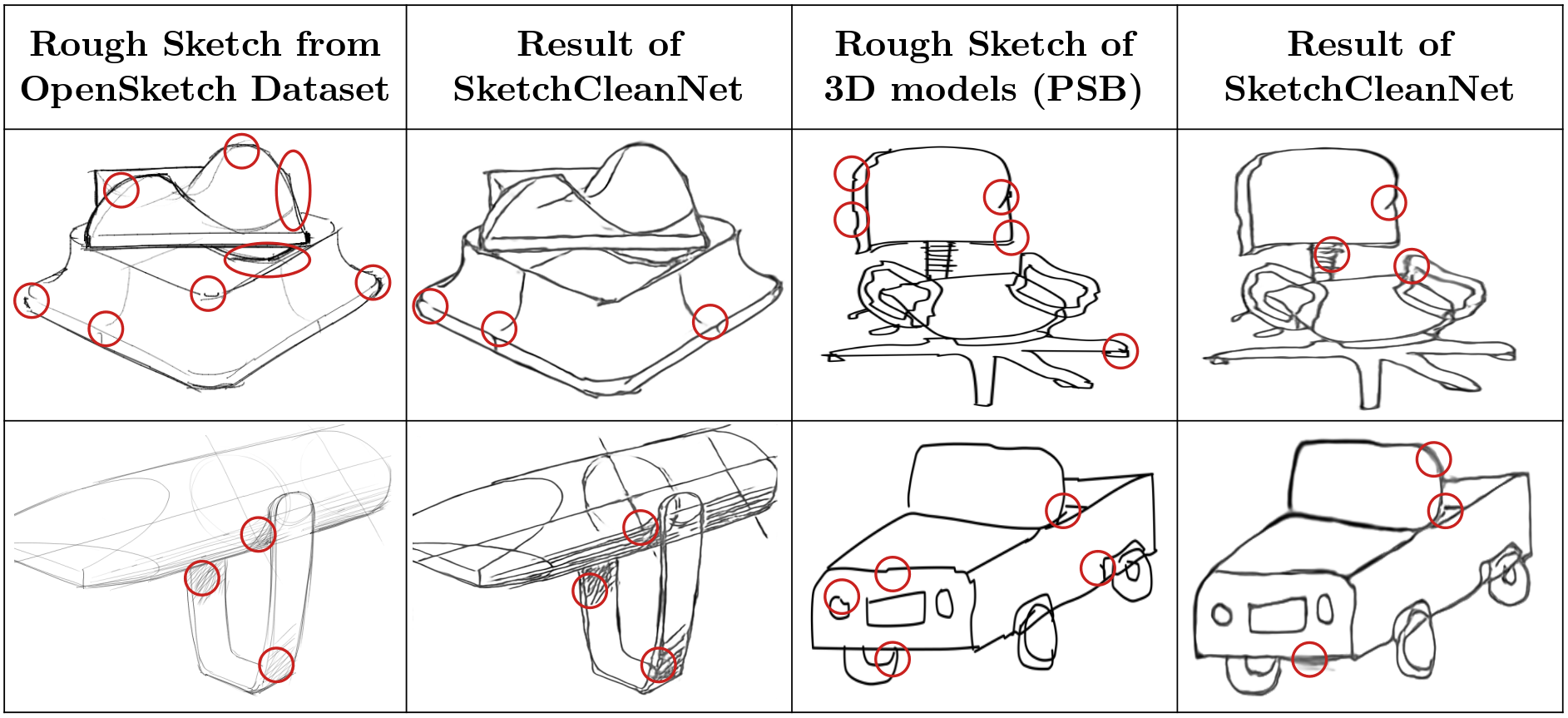}
\caption{Results of SCNet on sketches from OpenSketch and models of PSB. Results on OpenSketch: Image1 - simplification of multi-stroke regions; Image2 - no correction as multiple strokes are being treated as distinct input lines. Results on sketches of PSB: Unwanted blurring of some portions of the images since the input sketches already contain clear, sharp stokes.}
\label{fig:result_psb_open}
\end{figure*}

\subsection{Results of SCNet on other sketch datasets}
The performance of SCNet is also analysed on inputs from other sketch datasets such as OpenSketch \cite{OpenSketch19} and the sketches \textcolor{black}{corresponding to the} 3D shapes \textcolor{black}{of common objects} from the Princeton Shape Benchmark \cite{Eitz}. It is to be noted here that these datasets contain sketches of generic 3D shapes and not of 3D engineering parts and components. This experiment, therefore, serves as a cross-domain test of the performance of SCNet. Figure \ref{fig:result_psb_open} shows some sample results. 

For sketches from the OpenSketch dataset, minor defect corrections such as completion of missing lines, removing unwanted lines, etc. are achieved. The first sample result from Figure \ref{fig:result_psb_open} shows that certain multi-stroke portions are simplified. However, most sketches from the OpenSketch dataset contain images with heavy multi-stroke regions, and no improvement is observed for such images. This is probably because the network treats these lines as separate input lines/curves rather than multiple strokes for the same line/curve (see sample result in the second row). 

Most sketches of the PSB models are already simplified, have continuous lines and curves in the input, and do not require any major corrections. The results shown in Figure \ref{fig:result_psb_open} show only minor improvements, such as correction of zig-zag lines to uniform curves etc. However, there is some unwanted blurring of some portions of the image, as seen in both the sample results. This is because the PSB sketches already contain clear and sharp input lines / curves to begin with. Using SCNet, which is trained on rough sketch inputs with multiple defects \textcolor{black}{(specific to engineering components and not regular 3D shapes)}, might be unnecessary for such cases.

\subsection{Quantitative comparison of results by using enhanced query sketches for retrieval}
The ultimate test for evaluating the outputs generated by SCNet is done by actually verifying if the cleaned sketches provide better retrieval results as compared to the sketches with defects. A deep learning based search engine for 3D CAD models has been presented along with the CADSketchNet dataset by \cite{CADSketchNet}, which is modelled using a Siamese Network. It is also reported there, that a Siamese Network using ResNet18 for both CNN pipelines yielded the best results. A search engine with the same configuration is used in this paper as well, to compare the retrieval results when trained using (1) sketches from CADSketchNet and (2)  SCNet.

The results of retrieval using both the sets of query sketches are reported in Table \ref{tab:ret_compare}. For search and retrieval, the value for recall is simply the probability that a relevant data item is retrieved by the query, while precision denotes what fraction of the retrieved data items are relevant to the query. Clearly, higher values of precision and recall indicate better retrieval performance. In addition, a comparison of the top $k$-accuracy value is also reported. If 8 out of the 10 most similar results belong to the same class as the query, then the top $10$-accuracy value is 80\%. In Table \ref{tab:ret_compare}, the values reported against these metrics are the average values obtained over all categories of 3D CAD models in the database. These values show that using the cleaned sketches generated by SCNet gives better retrieval performance, for each of the evaluation metrics. This demonstrates that the retrieved results of a search engine are only as good as the input query, and also justifies the need for a dedicated query sketch enhancement module in an image-based search engine.

\section{Limitations and Future Work}
\label{sec:limi}
The current work focuses on the enhancement and correction of query sketches, aimed at improving the retrieval performance of a 3D CAD model search engine. Hence, \textcolor{black}{the network is trained specifically on sketches of engineering shapes and parts. Due to this, the network will not always translate well to sketches from other domains, such as the sketches of regular shapes (3D models of common objects).  Obtaining sketches from multiple contexts and augmenting the existing dataset to build unified model that works across domains is an interesting future work.}

The primary goal of the paper is to provide better query images for a CAD model search engine, and that goal has been achieved. However, a detailed analysis and characterization of the various types of defects present in the query sketches, and developing image enhancement solutions specific to the type of defect is worthy of further investigation. \textcolor{black}{Augmenting the dataset with sketches that contain varying degrees of noise and defects is also an interesting future work that helps in building a robust network.} \textcolor{black}{In addition, the current approach only considers query sketches from a single orientation of the 3D model. This is because the end-user of a search engine would typically expect a search result using a single query image. However, the proposed methodology could also be extended to process multi-view sketch queries, i.e, sketches of a CAD model from multiple orientations as opposed to a single query sketch.}

\section{Conclusion}
\label{sec:conc}
The retrieved results of a search engine are only as good as the input query. Typical query sketches of 3D CAD models contain various types of defects, such as missing portions, duplicate lines, the presence of unwanted mesh lines, etc. This paper, therefore, presents SketchCleanNet - a deep learning approach to enhance and correct such rough query sketches of 3D CAD models. The network is trained using a ground truth dataset of clean query sketches, and this dataset will be made available publicly along with the enhanced query images generated by the network. A detailed comparison of the results of SCNet with other architectures is reported, with the proposed architecture significantly outperforming other approaches. Input sketches from sketch datasets of generic 3D shapes are also used to test the performance of the network. Finally, the generated clean sketches from the network are then used as query images to train a deep learning based search engine for 3D CAD models. A quantitative comparison of retrieval results demonstrates that utilizing the query images from SCNet yields significantly better results for retrieval, while also justifying the need for a dedicated query sketch enhancement module in an image-based search engine.

\section*{Acknowledgments}
Thanks are due to the teams of the ESB and the MCB datasets for making their data publicly available. Thanks are also due to many users who have contributed to our CADSketchNet dataset.

\footnotesize
\bibliographystyle{unsrt}
\bibliography{refs}
\end{document}